\documentclass[conference]{IEEEtran}
\IEEEoverridecommandlockouts
\usepackage{cite}
\usepackage{amsmath,amssymb,amsfonts}
\usepackage{algorithmic}
\usepackage{graphicx}
\usepackage{textcomp}
\usepackage{xcolor}
\def\BibTeX{{\rm B\kern-.05em{\sc i\kern-.025em b}\kern-.08em
    T\kern-.1667em\lower.7ex\hbox{E}\kern-.125emX}}
\begin{document}

\title{A Comparison of CNN and Classic Features for Image Retrieval\\
}

\author{\IEEEauthorblockN{Umut \"Ozayd\i n}
\IEEEauthorblockA{\textit{LIACS} \\
\textit{Leiden University} \\
Leiden, the Netherlands \\
ozaydinumut@gmail.com
}
\and
\IEEEauthorblockN{Theodoros Georgiou}
\IEEEauthorblockA{\textit{LIACS} \\
\textit{Leiden University} \\
Leiden, the Netherlands \\
t.k.georgiou@liacs.leidenuniv.nl}
\and
\IEEEauthorblockN{Dr. Michael Lew}
\IEEEauthorblockA{\textit{LIACS} \\
\textit{Leiden University} \\
Leiden, the Netherlands \\
mlew@liacs.nl}
}


\maketitle

\IEEEpubidadjcol

\begin{abstract}
Feature detectors and descriptors have been successfully used for various computer vision tasks, such as video object tracking and content-based image retrieval. Many methods use image gradients in different stages of the detection-description pipeline to describe local image structures. Recently, some, or all, of these stages have been replaced by convolutional neural networks (CNNs), in order to increase their performance. A detector is defined as a selection problem, which makes it more challenging to implement as a CNN. They are therefore generally defined as regressors, converting input images to score maps and keypoints can be selected with non-maximum suppression. This paper discusses and compares several recent methods that use CNNs for keypoint detection. Experiments are performed both on the CNN based approaches, as well as a selection of conventional methods. In addition to qualitative measures defined on keypoints and descriptors, the bag-of-words (BoW) model is used to implement an image retrieval application, in order to determine how the methods perform in practice. The results show that each type of features are best in different contexts.
\end{abstract}

\begin{IEEEkeywords}
neural networks, keypoints, detectors, descriptors
\end{IEEEkeywords}

\section{Introduction}
As billions of images and millions of hours of video are uploaded on the Internet every year, the ability to decode their contents is a challenge for computers that has great value. However, extracting high-level information from raw-pixel data has been the topic of a lot of research. Most traditional methods rely on detection and extraction of salient features. A plethora of methods that detect and describe features have been proposed \cite{salahat2017recent,li2015survey}. One of the most popular methods is SIFT \cite{lowe2004distinctive}. Some methods approximate image gradients, as directly calculating them is an expensive operation and the added performance of calculating exact gradients is usually not worth the extra processing time \cite{bay2008speeded}. Other methods work by directly comparing pixel intensities and do not approximate gradients \cite{Rublee:2011:OEA:2355573.2356268,Leutenegger:2011:BBR:2355573.2356277,alahi2012freak}. Such methods may be preferable in on-line applications computational efficiency is critical.\par  
In recent years, researchers have proposed CNN based methods for detecting and extracting features \cite{lenc2016learning,verdie2015,savinov2017quad,yi2016,moo2016learning}. CNNs have been successfully used to solve complex problems, such as image classification or speech recognition, but their application to feature detection and description is relatively new. The learning process of these methods can generally be divided into two classes, \emph{supervised} learning and \emph{unsupervised} learning. They key difference between the two is existence of annotated data, which help the supervised methods have a very efficient learning process. The result though might be biased according to biases in the data labels. On the other hand unsupervised methods rely on the method's intrinsic ability to find clusters of similar data. This, however, makes the task more challenging and requires a specifically designed algorithm for learning. The goal of this paper is to review how CNNs can be used for feature detection and extraction and compare the performance of the resulting CNNs to the traditional, hand crafted, detectors and extractors.\par
Many works have compared different local feature detectors, either by comparing literature to their own method or making a generic comparison \cite{lenc2016learning,verdie2015,savinov2017quad,yi2016,moo2016learning,li2015survey,salahat2017recent}. To the best of our knowledge, there is minimal work that compared CNN based detectors. Most survey papers compare traditional methods \cite{li2015survey,salahat2017recent}, whilst papers that introduce CNN based detectors compare their own method to one or more traditional approaches \cite{lenc2016learning,verdie2015,savinov2017quad,yi2016,moo2016learning}. This paper performs an in depth comparison between CNN based and traditional detectors. We evaluate the methods based on robustness in the presence of common distortions (such as rotation, scaling, and noise) and also directly on the domain of image retrieval.\par

\section{Hand Crafted Detectors}
\label{sec:hand_crafted}
\begin{table*}[htbp]
\caption{CNN local feature detection methods}
\begin{center}
\begin{tabular}{|l|c|c|c|c|}
\hline
\textbf{Method} & \textbf{\textit{Supervision}} & \textbf{\textit{Detection}} & \textbf{\textit{Orientation Invariant}}& \textbf{\textit{Description}} \\
\hline
Covariant feature detector \cite{lenc2016learning}& covariant constraint & Yes & Depends on defined transformations & No \\
Quad detector \cite{savinov2017quad}& rank constraint & Yes & Depends on defined transformations & No \\
Orientation Estimator Network \cite{moo2016learning}& rotation loss & No & Yes & No \\
LIFT \cite{yi2016}& SIFT, rotation loss & Yes & Yes & Yes \\
\hline
\end{tabular}
\label{tab:cnnBasedMethods}
\end{center}
\end{table*}
As mentioned above, a plethora of different hand-crafted local feature detectors have been proposed in the literature. Due to space limitations , we selected a representative set from the literature \cite{li2015survey,salahat2017recent,liu2016extracting}. More specifically, we will utilize the well known \textbf{SIFT} \cite{lowe2004distinctive} and \textbf{SURF} \cite{bay2008speeded}, \textbf{CenSureE} \cite{Agrawal2008}, a detector leveraging the computational efficiency of SURF's box filters and the accuracy of SIFT's circular filters and \textbf{AGAST} \cite{mair2010}, a more accurate and efficient extension of the FAST detector \cite{rosten2006}.\par

\section{cnn based detectors}
\label{sec:cnn_approaches}
In the past few years, several CNN approaches that try to detect keypoints have been proposed. The first, main characteristic these methods can be divided by is whether the training objective depends on predefined keypoints (e.g. SIFT keypoints) or they learn in an "unsupervised" manner. In the first category the methods extract keypoints using an existing method, such as SIFT and train the network to predict where these points appear. This method has the benefit of using very clear supervised signals which result in an efficient training procedure. On the other hand, they are limited on the capabilities of the method they try to approximate. In the second category are methods that define a loss function that is minimized when the network is able to identify points that show certain characteristics. None of the methods presented is truly unsupervised, since a set of transformations is defined and a very specific dataset tailored to the objective function created in order to train the networks. Moreover, there are several features that a method may or may not have, such as orientation assignment and whether it provides a description method. All the methods are briefly discussed below and summarized in Table \ref{tab:cnnBasedMethods}. Many researchers have used the activations of pre-trained VGG network on the ImageNet 2012 dataset, as descriptors for keypoints \cite{tolias2015}. Thus, we also try to combine them with some of the detectors.\par
\textbf{Covariant Feature Detector} (CFD) \cite{lenc2016learning}: With CFD, a network is supervised by a covariant constraint and is forced to resolve a set of transformations while be covariant to all transformations defined.\par
\textbf{Quad Detector} \cite{savinov2017quad}: It is called quad-detector because it learns from quadruples of input patches. The output of the network is supposed to be a heat map showing features at the maximum. Given two points and a transformation of those, the network is asked to keep the ranking of the points before and after the transformation. The invariance of the network is handled by the train set, in which all the desired transformations are defined. The transformations can either be included in the dataset, or manually applied.\par
\textbf{LIFT} \cite{yi2016}: LIFT is the only method proposed that includes detection, explicit orientation estimation and description. Overall it defines three networks, one for each task. The LIFT can be also viewed as a combination, with slight adjustments, of the TILDE \cite{verdie2015} and the Orientation Estimation Network (OEN)\cite{moo2016learning}. TILDE is a network that learns to identify patches that are centered around a keypoint. This is achieved through a classification loss, as well as two regularization losses. One to force local maxima of the output have higher peaks and the other helps identify the same keypoints under different image conditions. OEN on the other hand, does not detect keypoints, but given a keypoint it assigns it an orientation. Given an image patch, the network outputs two values, which are interpreted as scaled the sine and cosine of the angle patch.The network tries to minimize the difference between the description of rotated patches.\par
The LIFT pipeline works as following. Initially the descriptor network is trained, where its given pairs of image patches around SIFT keypoints. It learns to produce a descriptor which is similar for matching pairs and dissimilar for pairs of different keypoints. Then the orientation network is trained similarly to OEN, coupled with the already pre-trained descriptor. Lastly, the detector is trained in a similar fashion to TILDE where it tries to classify SIFT keypoints. The loss function is different than TILDE's, as it leverages quadruples of patches and forces the detector to favor points that are invariant to the transformations of the training data.

\section{Experimental setup}
\label{sec:experimental_setup}

There are a numerous metrics that measure the quality of detectors \cite{salahat2017recent}, repeatability being the most frequently used. Moreover, we measure the descriptor matching precision and recall, as well as the non-redundancy and coverage of the key points produced as defined in \cite{ehsan2011,rey-otero2015}. Besides the theoretical characteristics an ideal detector needs to have, it mostly needs to provide points that lead to good performance on real life applications. Thus, we test the detectors both on theoretical measurements (i.e. repeatability) and on a real world application, i.e. Image Retrieval (IR).\par
\begin{figure*}
  \caption{The repeatability scores of the manual transformations and the lighting changes.}
  \begin{tabular}{@{}c@{}c@{}}
    \includegraphics[width=.45\linewidth]{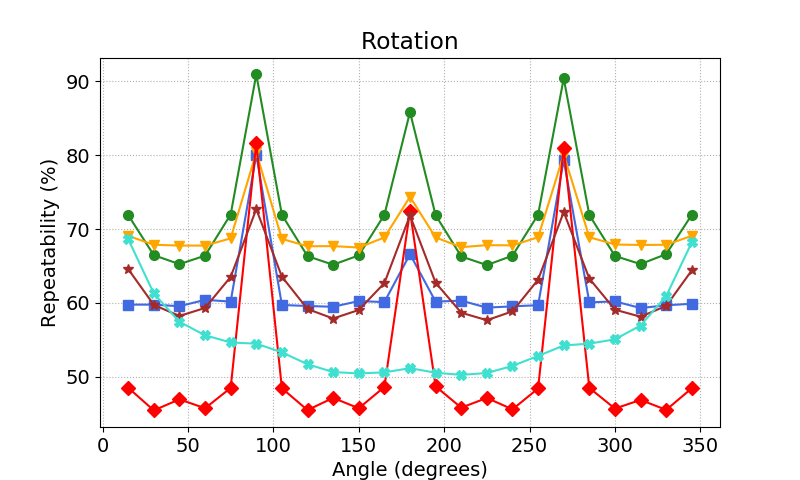} &
    \includegraphics[width=.45\linewidth]{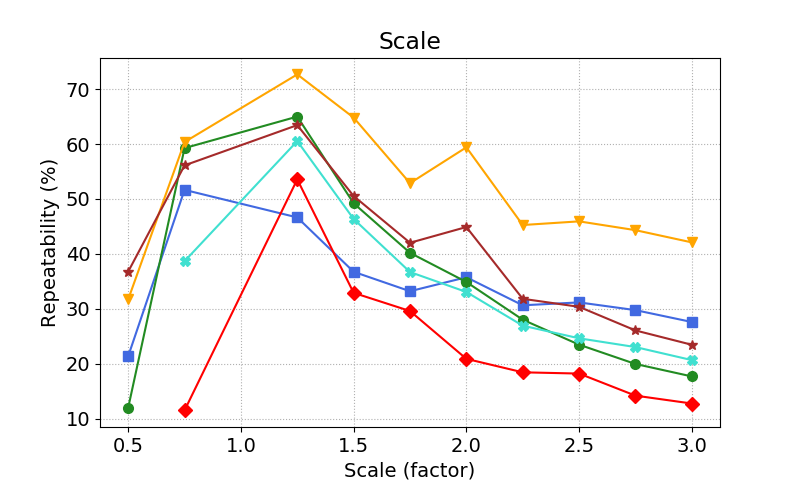} \\[\abovecaptionskip]
    \includegraphics[width=.45\linewidth]{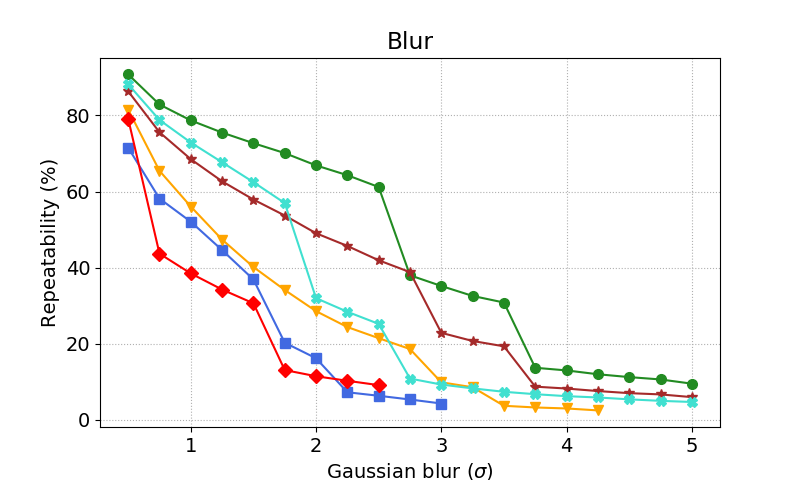} &
    \includegraphics[width=.45\linewidth]{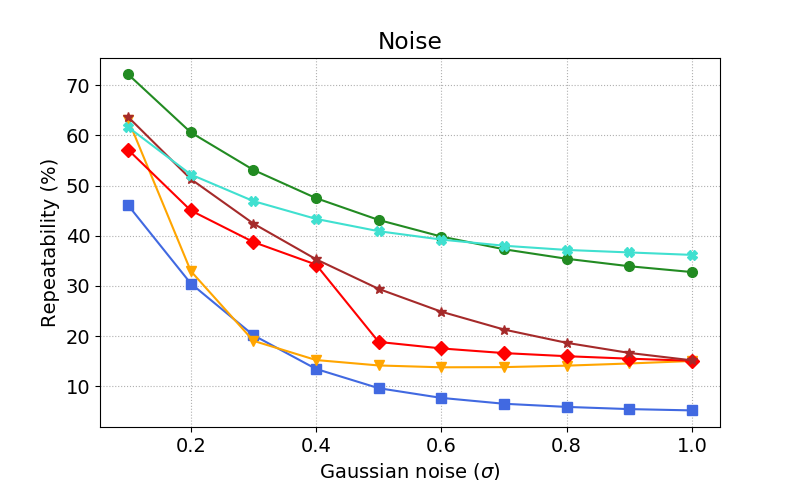}\\[\abovecaptionskip]
    \multicolumn{2}{c}{\includegraphics[width=\linewidth]{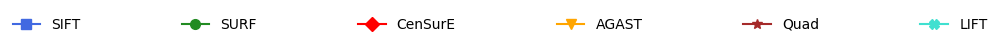}}
  \end{tabular}
  \label{fig:repeatability_results}
  \end{figure*}

For the repeatability experiment, we chose the first $2$K images from the test set of the ImageNet 2017 set. On every image, we perform rotation, scaling, blur and add Gaussian noise. We measure the repeatability for different levels of these transformations. In order to make a fair comparison, the $K \in \{300,600,1000\}$ most prominent keypoints from each detectors were used. The results in Figure \ref{fig:repeatability_results} show the average performance over the three $K$ values.\par

  \begin{figure*}
  \caption{The matching scores of the detectors using all transformations.}
  \begin{tabular}{@{}c@{}c@{}}
    \includegraphics[width=.45\linewidth]{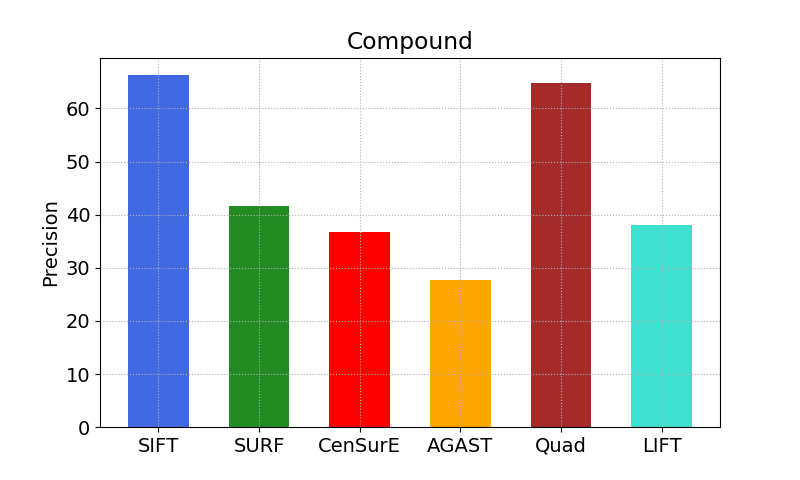} &
    \includegraphics[width=.45\linewidth]{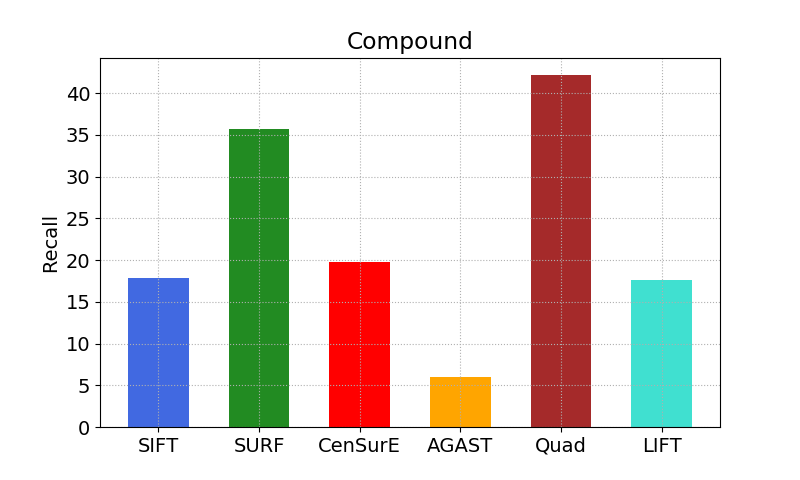} \\[\abovecaptionskip]
  \end{tabular}
  \label{fig:matching_results}
  \end{figure*}

  \begin{figure*}
  \caption{The non-redundancy and coverage scores of the detectors.}
  \begin{tabular}{@{}c@{}c@{}}
    \includegraphics[width=.5\linewidth]{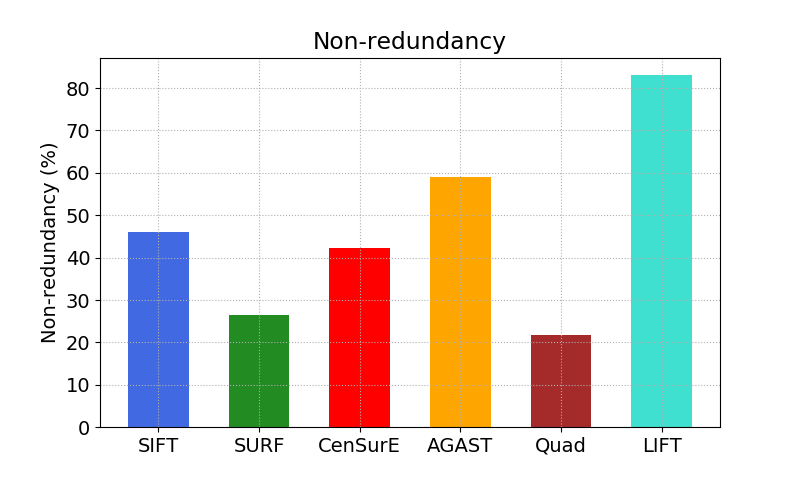} &
    \includegraphics[width=.5\linewidth]{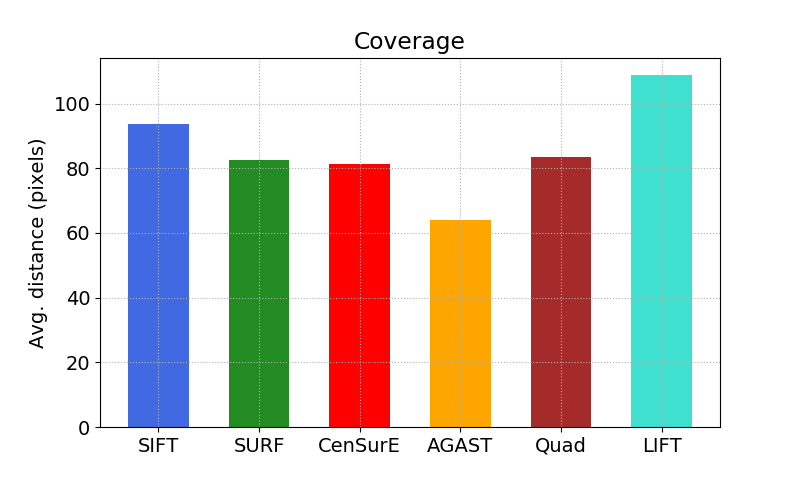} \\[\abovecaptionskip]
  \end{tabular}
  \label{fig:coverage_results}
  \end{figure*}

  \begin{table}[htbp]
  \caption{The combinations of detector, orientation estimator and descriptor used.}
  \label{tab:retrieval-combis}
  \begin{tabular}{|c||c|c|c|}
  \hline
  \textbf{Method} & \textbf{Detector} & \textbf{Or. Estimator} & \textbf{Descriptor}\\
  \hline
  SIFT \cite{lowe2004distinctive} & SIFT & SIFT & SIFT\\
  \hline
  SIFT + VGG16 \cite{lowe2004distinctive, simonyan2014very} & SIFT & SIFT & VGG16\\
  \hline
  SURF \cite{bay2008speeded} & SURF & SURF & SURF\\
  \hline
  CenSurE \cite{Agrawal2008} & CenSurE & SURF & SIFT\\
  \hline
  AGAST \cite{mair2010, Leutenegger:2011:BBR:2355573.2356277} & AGAST & BRISK & BRISK\\
  \hline
  Quad \cite{savinov2017quad} & Quad & SIFT & SIFT\\
  \hline
  Quad + VGG16 \cite{savinov2017quad, simonyan2014very} & Quad & SIFT & VGG16\\
  \hline
  LIFT \cite{yi2016} & LIFT & SIFT/LIFT & LIFT\\
  \hline
  \end{tabular}
  \end{table}
 
In real-world image retrieval systems, the database may contain millions of images. While the comparison of every pair of descriptors is a very precise method, it is too costly to perform with these numbers of images. One method to achieve real-time performance is the \emph{bag-of-words} (BoW) model \cite{zheng2018sift, sivic2003}, which is used in these experiments. The BoW trainer and histogram extractor implemented in the OpenCV library are used. The number of database images is set to $250K$ and the number of clusters to $8K$. The images are taken from the ILSVRC 2012 train set. The dictionary is calculated by taking the descriptors of the 10 most prominent keypoints in each image, resulting in $2.5M$ descriptors. For the histograms, $200$ keypoints per image are used. In the final step, $2.5K$ query images are constructed by applying a random scale, rotation, blur, noise, contrast and brightness. The query images are then compared to the database, using the L1-distance, and all database images are ranked by their distance to the query image. If the highest ranked image is the same as the untransformed query image the result is considered correct. As quality measurements we use the accuracy (zero rank) over all test query images and the average rank of the correct image.\par

\begin{table*}[htbp]
  \centering
  \caption{The accuracy (top-1) and average rank of the image retrieval benchmark.}
  \label{tab:res_retrieval}
  \begin{tabular}{|c||c|c|c|c|c|c|c|c|}
  \hline
  Method & SIFT & SIFT+VGG16 & SURF & CenSure & AGAST & Quad & Quad + VGG16 & LIFT \\
  \hline
  Correct (\%) & 43.12 & 24.24 & \textbf{60.16} & 48.48 & 10.64 & 59.52 & 36.48 & 8.08 \\
  \hline
  Avg. rank & 4133 & 10121 & \textbf{1180} & 2288 & 25980 & 1544 & 4914 & 24718 \\
  \hline
  \end{tabular}
  \end{table*}
\section{Results}
\label{sec:experiments}
All the experimental setup was implemented in C++ and Python. The OpenCV 3 implementations were used for all conventional features detectors. The parameters of each method were set to values as close as possible to those given in the original papers, and otherwise to the standard values present in the library. The combinations of detectors and descriptors used for the experiments can be seen in Table \ref{tab:retrieval-combis}.\par

The Quad-Detector was re-implemented by the authors based on the original paper in PyTorch, and trained on the DTU Robot Image Point Feature Data Set \cite{aanaes2011}, using only illumination changes as correspondences. For the covariant feature detector, scale-space and orientation assignment of SIFT were used, but it did not result in stable keypoints and thus omitted from the experiments. For the VGG descriptor we used the 7th layer of TensorFlow's \cite{tensorflow2015-whitepaper} VGG-16 implementation. The input to the network was a patch 40x40 resulting in an output vector 1x1x256. Finally, for the LIFT detector two implementations are used. The first one, tested on repeatability did not use large rotations to train the rotation estimator, and thus the SIFT orientation assignment was used. The second version which train the rotation estimator also on large image rotations was used  in the IR experiment.\par

Figures \ref{fig:repeatability_results}, \ref{fig:matching_results} and \ref{fig:coverage_results} show the results of the repeatability, matching and non-redundancy/coverage experiments respectively. Finally, Table \ref{tab:res_retrieval} shows the percentage of images that were retrieved correctly, as well as the average rank of all the $2.5$K instances.\par
The method that has the highest performance is SURF, as it has the most correct matches and the lowest average rank. Moreover, it consistently has top perfrormance on the repeatability experiments. SIFT did not perform as well. This may partially result from the fact that the transformations contain blurring and noise, which SIFT is sensitive to. AGAST performed very well on the repeatability under rotation and scaling, but had poor performance on the rest of the experiments. CenSurE also worked relatively well in IR, with 48\% of the predictions correct, but had poor performance on the repeatability experiments.\par
From the deep learning methods, the Quad detector had the highest performance in the IR results whilst in repeatability LIFT performed better in all measurements except rotations. An interesting point is that although SIFT and Quad have only the detectors different, the IR performance is very different.\par

\section{Conclusions}
\label{sec:conclusions}
CNNs have managed to outperform traditional methods in many applications. Our experiments show that when it comes to local feature detection they perform competitively in certain contexts. From the handcrafted methods, SURF had the best overall performance and was clearly competitive with the top CNN approaches.\par
For the different CNN strategies, we can see that using SIFT points (see \cite{yi2016}) to supervise the learning procedure did not raise competitive performance in a real life application, which can be explained by the fact the we are trying to learn an imperfect method. On the other hand, LIFT managed to get competitive performance in repeatability with the presence of blur and noise. The more "unsupervised" method managed to produce one of the best results, in most of our experiments. Nonetheless, it underperformed in the presence of rotation. Finally, our experiments show that using VGG features instead of a handcrafted descriptor in the pipeline gives lower performance. Still, more experiments would be required to make a conclusive remark, as there are numerous ways to utilize a pre-trained network for that purpose.\par
Our viewpoint is that many researchers (including some in our group) have come to a default assumption that deep learned approaches will greatly outperform the classic handcrafted features.  This paper shows that it depends on the exact problem(s) one is trying to solve.  For example, rotation is a major challenge for the CNN based methods and more generally it is clear that additional research is warranted before turning away from the handcrafted features.
\bibliographystyle{IEEEtran}
\bibliography{refs.bib}
\end{document}